\documentclass[twoside,11pt]{article}

\usepackage{jmlr2e}
\usepackage{verbatim}
\usepackage{epstopdf}
\usepackage{subfig}
\newcommand{\dee}[0]{\, \mathrm{d}}

\newcommand{\avg}[1]{\langle #1 \rangle}

\usepackage{amsfonts}
\usepackage{amsmath}

\ShortHeadings{Polynomial expansion of the binary classification function}{P\'eter K\"oves\'arki}
\firstpageno{1}

\begin{document}
\title{Polynomial expansion of the binary classification function}
\author{\name P\'eter K\"oves\'arki \email kovesarki@physik.uni-bonn.de \\
	\addr Physikalisches Institute \\ Universit\"at Bonn \\
		Bonn, 12 Nussallee, D-53115 DE }
\editor{}

\maketitle

\begin{abstract}%
This paper describes a novel method to approximate the polynomial coefficients of regression functions, with particular interest on multi-dimensional classification. The derivation is simple, and offers a fast, robust classification technique that is resistant to over-fitting. 
\end{abstract}


\section{Procedure to calculate the classification polynomial}

The goal of binary classification is to find the distinction between a signal $s(x)$ and background $b(x)$ probability distribution.  The optimal separation contours are described by \cite{Neyman01011933}, and it is well known these contours can be found by binomial regression \citep[see][]{Bishop:2006:PRM:1162264}.  In neural networks it is typically a regression between target values $\pm 1$, and the optimal response function $F(x)$ is related to the $P(s|x)$ purity of signal:

$$ F(x) = \frac{s(x) - b(x)}{s(x)+b(x)} = 2P(s|x) -1 \,.$$
By reordering, performing a Fourier transformation and using the Taylor expansion of F(x), it becomes 

%
\begin{equation} \label{eq:ReorderedFTdF}
 \sum_{k=0}^{\infty} \frac{1}{k!}F^k \int_\mathbb{R}  x^k \left( s(x) + b(x) \right) e^{i\omega x} \dee x = \int_\mathbb{R}  \left( s(x) - b(x) \right) e^{i\omega x} \dee x \,.
\end{equation}
The Taylor series of characteristic functions can be expressed with the $\avg{x^k}$ moments of the corresponding distribution, which is used in the following definition of the $g(x)$ and $h(x)$ functions and their Fourier transforms $\hat g(\omega)$ and $\hat h(\omega)$:

$ g(x) := s(x) + b(x)$
$$ \hat g(\omega) = \sum_{k=0}^{\infty} i^{-k}\underbrace{\left( \avg{x^k}_s + \avg{x^k}_b \right)}_{\hat g^k} \frac{\omega^k}{k!}$$
$$ \frac{\partial^j}{\partial \omega^j}g(\omega) = \sum_{k=0}^{\infty} i^{-k}  \frac{\omega^k}{k!} \cdot \hat g^{k+j}  $$

$ h(x) := s(x) - b(x)$
$$ \hat h(\omega) = \sum_{k=0}^{\infty} i^{-k}\underbrace{\left( \avg{x^k}_s - \avg{x^k}_b \right)}_{\hat h^k} \frac{\omega^k}{k!} \,.$$
Substituting $g(x)$ and $h(x)$ back into eq.~\eqref{eq:ReorderedFTdF}, and exploiting that the Fourier transform of $x^kg(x)$ can be expressed with the $k$th derivative of $\hat g(\omega)$, one gets an equation that is true for every $\omega$, hence the equation should hold for the coefficients of $\omega^k$ for every $k$ as follows: 

$$ \sum_{k=0}^\infty i^{-k} \frac{\omega^k}{k!} \sum_{j=0}^{\infty} \frac{1}{j!}F^j \hat g^{k+j} = \sum_{k=0}^\infty i^{-k} \frac{\omega^k}{k!} \hat h^k$$
\begin{equation}\label{eq:HsumFG}
 \hat h^k = \sum_{j=0}^{\infty} \frac{1}{j!} F^{j} \hat g^{k+j}\,.
\end{equation}
These equations can be solved for $F^j$ either by a deconvolution, or by finding the solution for the matrix equation

\begin{equation} \label{eq:matrixHsumFG}
 \begin{pmatrix}  \hat h^0 \\ \hat h^1 \\ \vdots \\ \hat h^k \\ \vdots  \end{pmatrix} = 
	\begin{pmatrix} \hat g^0 & \hat g^1 & \cdots & \hat g^k \cdots \\  
				\hat g^1 & \hat g^2 & \cdots & \hat g^{k+1} \cdots \\
				\vdots & \vdots & 	\ddots & \vdots &  \\
				\hat g^k & \hat g^{k+1} & \cdots & \hat g^{2k} \cdots \\
				\vdots & \vdots & 	\ddots & \vdots &  	
				\end{pmatrix} 
	\begin{pmatrix}  F^0 \\  F^1 \\ \vdots \\ F^k \\ \vdots  \end{pmatrix}\,,
\end{equation}
where the ${}^1/_{k!}$ coefficients were suppressed into the $F^k$ unknowns, which also simplifies the later evaluation of the $F(x)$ function.
	
\begin{figure}[h]
   \subfloat[training sample]{\includegraphics[width=3in]{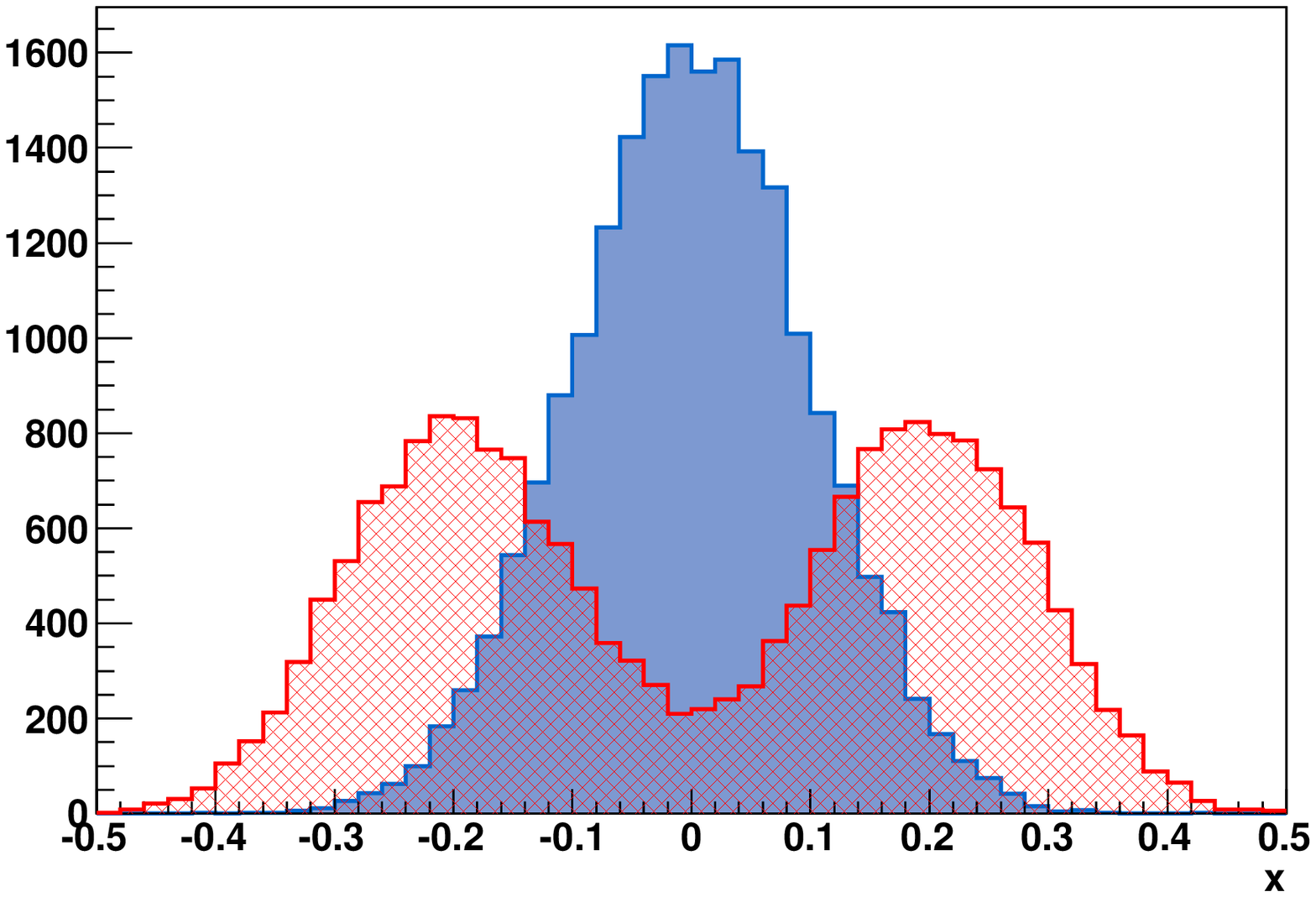}  	
   \label{fig:GaussSigBgr}}
  \subfloat[classification with polynomial]{\includegraphics[width=3in]{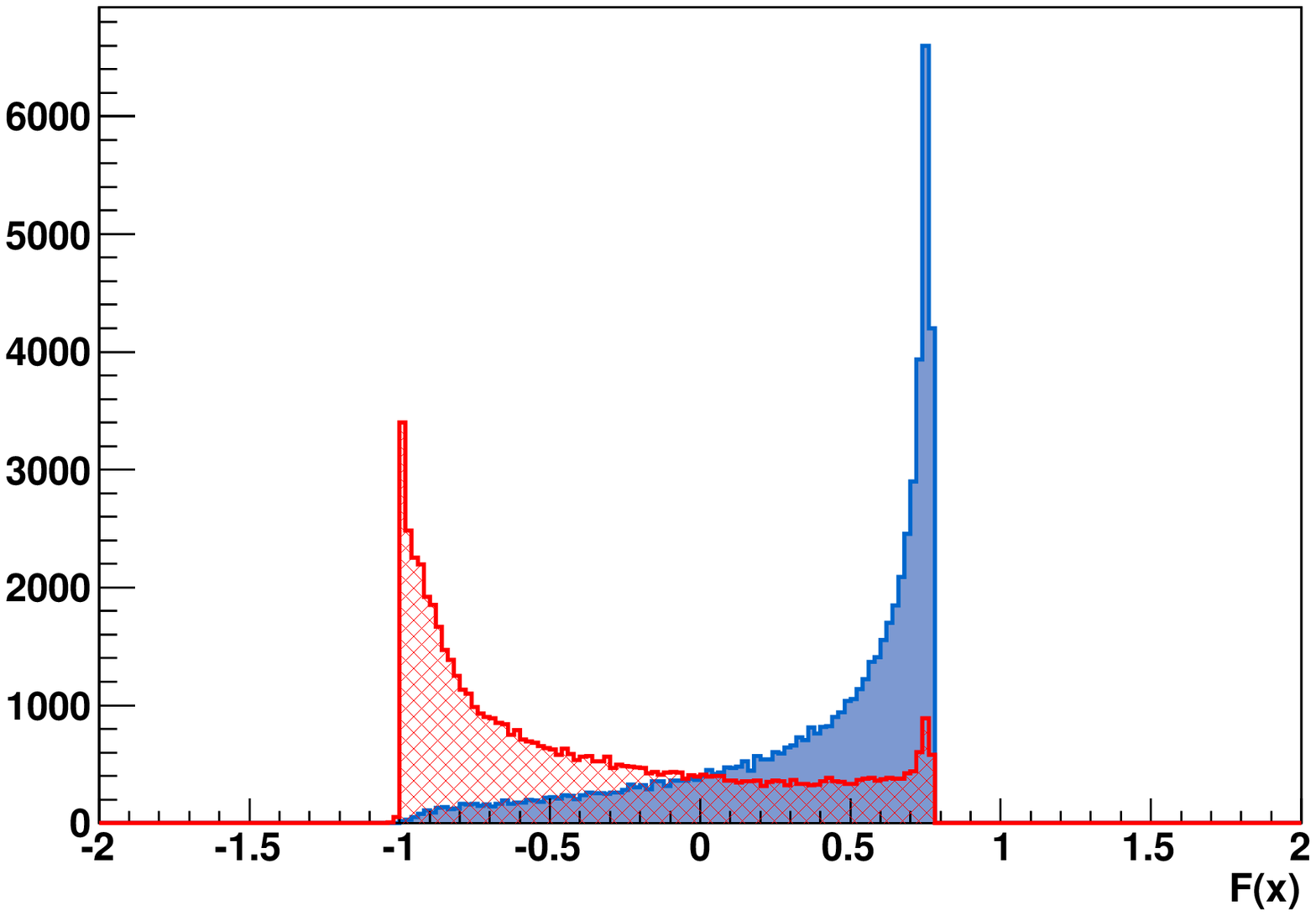} 
   \label{fig:GaussFx}} 
   \\
  \subfloat[monotonic prediction of the target]{ \includegraphics[width=3in]{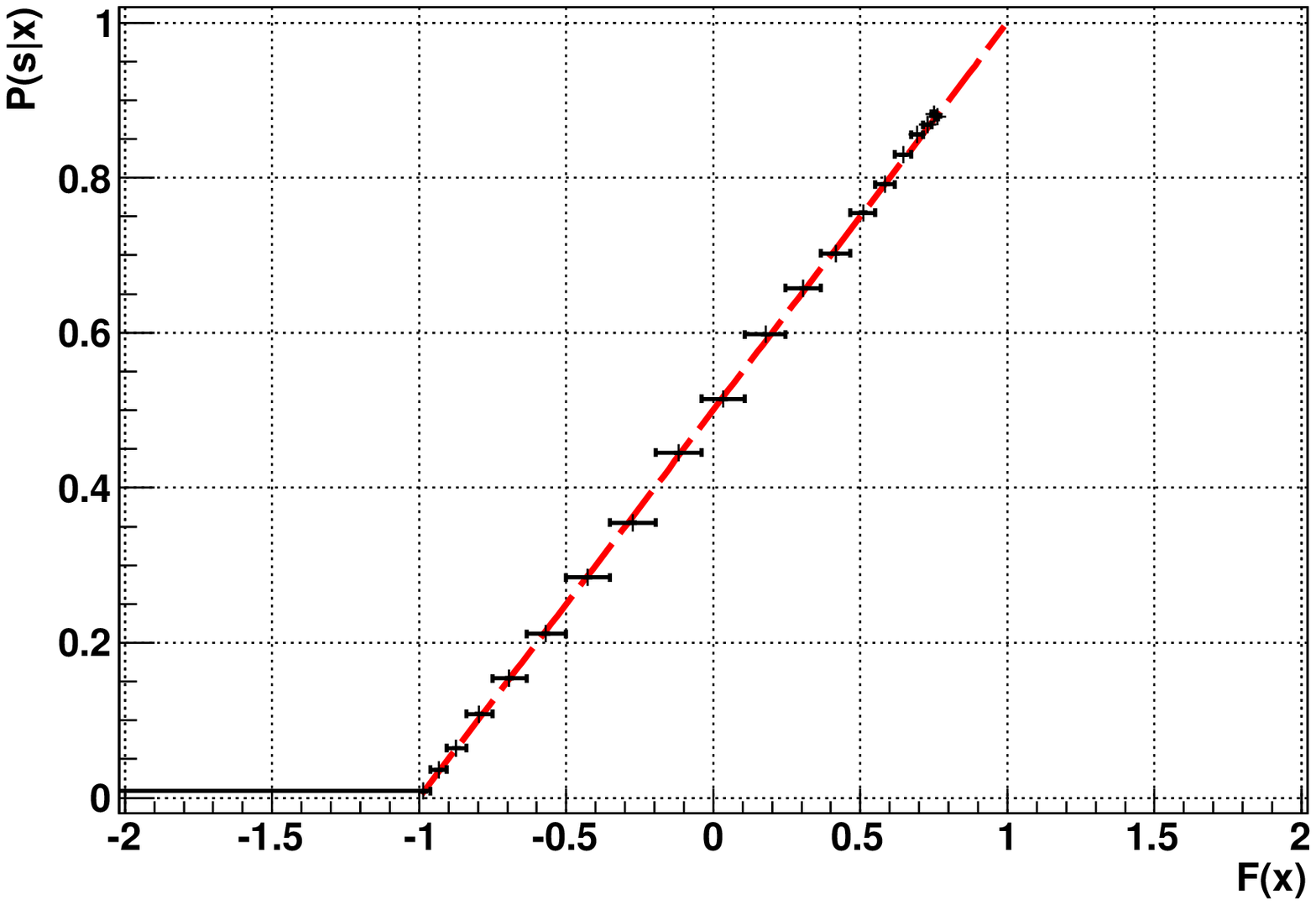} 
   \label{fig:GaussPurResp}}
   \subfloat[the theoretically optimal separation]{\includegraphics[width=3in]{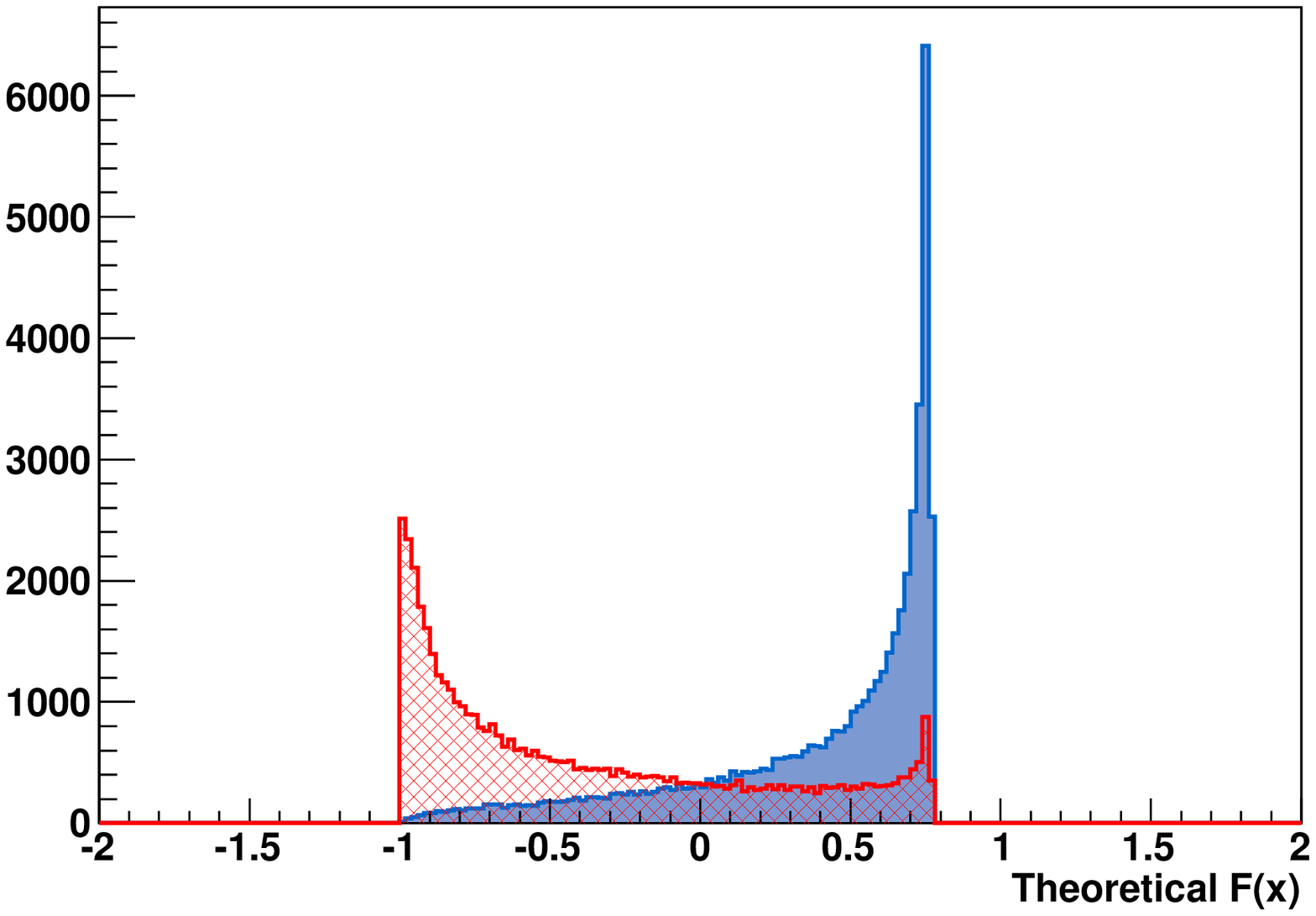} 
   \label{fig:GaussTheorFx}}
\caption{\small 
\ref{fig:GaussSigBgr} Example distribution with a Gaussian signal (solid blue) and a background of two Gaussian peaks (meshed red).
\ref{fig:GaussFx} Separation of signal from background with a 20 degree polynomial $F(x)$, comparable with the optimal separation 
on~\ref{fig:GaussTheorFx}. \ref{fig:GaussPurResp} The purity of the signal, evaluated for binned $F(x)$ values shows the linear dependency on $F(x)$, close to the ideal value (dashed red line).
}
\label{fig:1dimresults}
\end{figure}

A possible approximation is using the upper left $k\times k$ part of the matrix, and solve the finite system of equations. An example can be seen on fig.~\ref{fig:1dimresults}, where a 20 degree polynomial was used as a classifier on a Gaussian mixture sample with $10^4$ events, while the testing was done on an independent $10^4$ events. The resulting separation power is very similar to the theoretical optimum. Figure~\ref{fig:GaussPurResp} clearly shows, that the purity $P(s|x)$, evaluated in bins of $F(x)$ has a monotonic dependence on $F(x)$ itself. Lower order approximations of $F(x)$ may produce a non-linear, but still monotonically ascending curves, which feature is a requirement for a good classifier, as one can safely say that the events right to a certain $F(x)$ value are more signal like than the events to the left.

\section{Optimisations for multi-dimensional input}

In case the dimension of the input is greater than one, the $n$th moment of the distributions become $n$th order tensors, and similarly $\hat g^k$, $\hat h^k$ and $F^k$. The equation that connects these three tensor series is similar to eq.~\eqref{eq:HsumFG}, except that is has free tensor indices:

\begin{equation} \label{eq:TensorHsumFG}
 \hat h^k_{\mu_1 ... \mu_k } = \sum_{j=0}^{\infty} \hat g^{k+j}_{\mu_1... \mu_k \nu_1...  \nu_j } F^j_{\nu_1 ... \nu_j} \,.
 \end{equation}
Although this is a tensor equation, the indices of $\hat h^k_{\mu_1 ... \mu_k }$ and $F^j_{\nu_1 ... \nu_j}$ can be serialised, while $\hat g^{k+j}_{\mu_1... \mu_k \nu_1...  \nu_j }$ can be turned into a rectangular matrix with the serialised indices of ${\nu_1 ... \nu_j}$ as columns, and $\mu_1... \mu_k$ as rows. These system of equations can be rewritten as a block matrix equation, similar to eq.~\eqref{eq:matrixHsumFG}. However, a $d$-dimensional, $n$th order symmetric tensor has only  $ n + d -1 \choose n $ free parameters from the possible $d^n$. The difference can be many orders of magnitude even for small $d$ and $n$, therefore to speed up computations and use less memory, it is beneficiary to compactify the tensors in question, in a way described by \cite{Ballard6008988}.

For symmetric tensors with degree $n$, the component belonging to an index vector $\tilde \mu = \{\mu_1, ... ,\mu_n\}$ is the same for any permutation of $\mu_i$. These set of indices can be uniquely identified with the monomial of the index $m(\tilde \mu) = \{ m_1, ... ,m_d\}$, which is a $d$-dimensional vector, where $m_i$ is the multiplicity of the value $i$ in the index vector $\tilde \mu$. The multiplicity of a given monomial is the multinomial coefficient:

$$ \text{Multiplicity of } \{m_1, ... , m_d\} = {n \choose m_1, ... , m_d} = { n! \over m_1!  \cdots m_d!} \,.$$
The tensor multiplications in question, between the tensors $g^{k+j}_{\mu_1... \mu_k \nu_1...  \nu_j }$ and $F^j_{\nu_1 ... \nu_j}$,  can be simplified by running over only the free parameters, indexed with the monomials:

$$ g^{k+j}_{\mu_1... \mu_k \nu_1...  \nu_j } F^j_{\nu_1 ... \nu_j} = \sum_{m \in m(\nu_1, \cdots, \nu_j)} \hat g_{\mu_1...\mu_k m} F^j_m {n \choose m_1, ... , m_d} \,.$$
The multiplicity factor can be factored into $F^j_m$, just as the ${}^1/_{j!}$ terms before, with the benefit that the same terms should be used when $F^j_m$ is collapsed with the tensor of an input vector $x_{\nu_1}\cdots x_{\nu_j}$, in order to evaluate $F(x)$. The remaining indices of $\hat g^{k+j}_{\mu_1 \cdot \mu_k m}$ still hold a $k$-fold symmetry, just as $\hat h^k_{\mu_1 \cdot \mu_k}$, which can be simplified with the same procedure. In the eq.~\eqref{eq:TensorHsumFG}, there is no summation over the free indices of $\hat h^k$, hence there is no need for the multiplicity factors either. This makes it possible to create a large vector of the serialised $\hat h^k$ tensors, and a symmetric block matrix $G_{jk}$, containing the rectangular matrix version of $\hat g^{k+j}$ for every $j$ and $k$. 

The difficulty in creating this block matrix is, that despite most of the $\hat g^{k}$ tensors are used multiple times, they have to be partitioned into matrices in different ways. The efficient way of storing $\hat g^k$ in memory was described above, but to create the matrix versions one needs to access the tensor elements according to the simplified, monomial indices $m_k$  and $l_j$ of order $k$ and $j$ for the tensor $\hat g^{k+j}_{m_k, l_j}$ that matches with the structure of indices of $\hat h^k_{m_k}$ and $F^j_{l_j}$. In case the elements in any of the tensors above are stored serially in lexical index order,
 then for any index $\tilde \mu = \{\mu_1,\cdots, \mu_k\}$ it is true that $\mu_1 \leq \mu_2 \leq \cdots \leq \mu_k$. For the $\nu$ indices on the diagonal, where  $\nu_1 = \nu_{2} = \cdots = \nu_k$, it is possible to calculate the number of elements with higher lexically ordered index, because those indices map the free elements of a $d-\nu_1$ dimensional symmetric tensor of order $k$. The same way, the position of the generic $\mu$ index can be found by first finding the  $p_1$ position\footnote{In this paper it is assumed that the indexing starts from $\mu = \{1, \cdots, 1\}$, and the first position is $\text{pos}(\mu) = 1$} for the diagonal index $\{\mu_1+1, \cdots, \mu_1+1\}$, then calculating the $p_2$ position for the $\eta$ index, where $\eta_1 = \mu_1$, but $\eta_j = \mu_2+1$ for every $j>1$. It is done by simply subtracting from the $p_1$ position the number of free elements in a $d - \mu_2$ dimensional symmetric tensor of order $k-1$. Repeating this until the last element of the $\mu$ index, the formula to calculate its position reveals as


$$ \text {pos}(\mu) =  \underbrace{ k+d-1 \choose k }_{\text{No. free elements of a sym. tensor of order $k$} } - \sum_{i=1}^k {  \overbrace{(k-i+1)}^{\text{suborder}} + \overbrace{(d-\mu_i)}^{\text{subdimension}} -1 \choose k-i+1  } \,.$$

To match the elements of the serialised $\hat g^{k+j}_{o_{k+j}}$ tensor with the partitioned $\hat g^{k+j}_{m_k, l_j}$ matrix, one only has to combine the lexically ordered indices of $\hat g^{k+j}_{m_k, l_j}$ into one combined index with $k+j$ elements, which can be lexically ordered again with the help of its monomial. 

\section{Tests and conclusions}

The following example was made on a three dimensional sample, consisting of twelve non-overlapping, non-symmetric Gaussian peaks as signal and a flat background.
The classifier is a 20 degree multi-polynomial, which was found by solving a matrix equation with $1771 \times 1771$ elements in a few seconds with the solver of the Lapack library~\cite[see][]{laug}. Calculating the elements of this matrix from $2\times40$ thousand events takes about 20 seconds on a single core $2~\text{GHz}$ computer.

\begin{figure}[h]
	\centering
	\subfloat[multi-dimensional classification with polynomial]{
		\includegraphics[width = 3in]{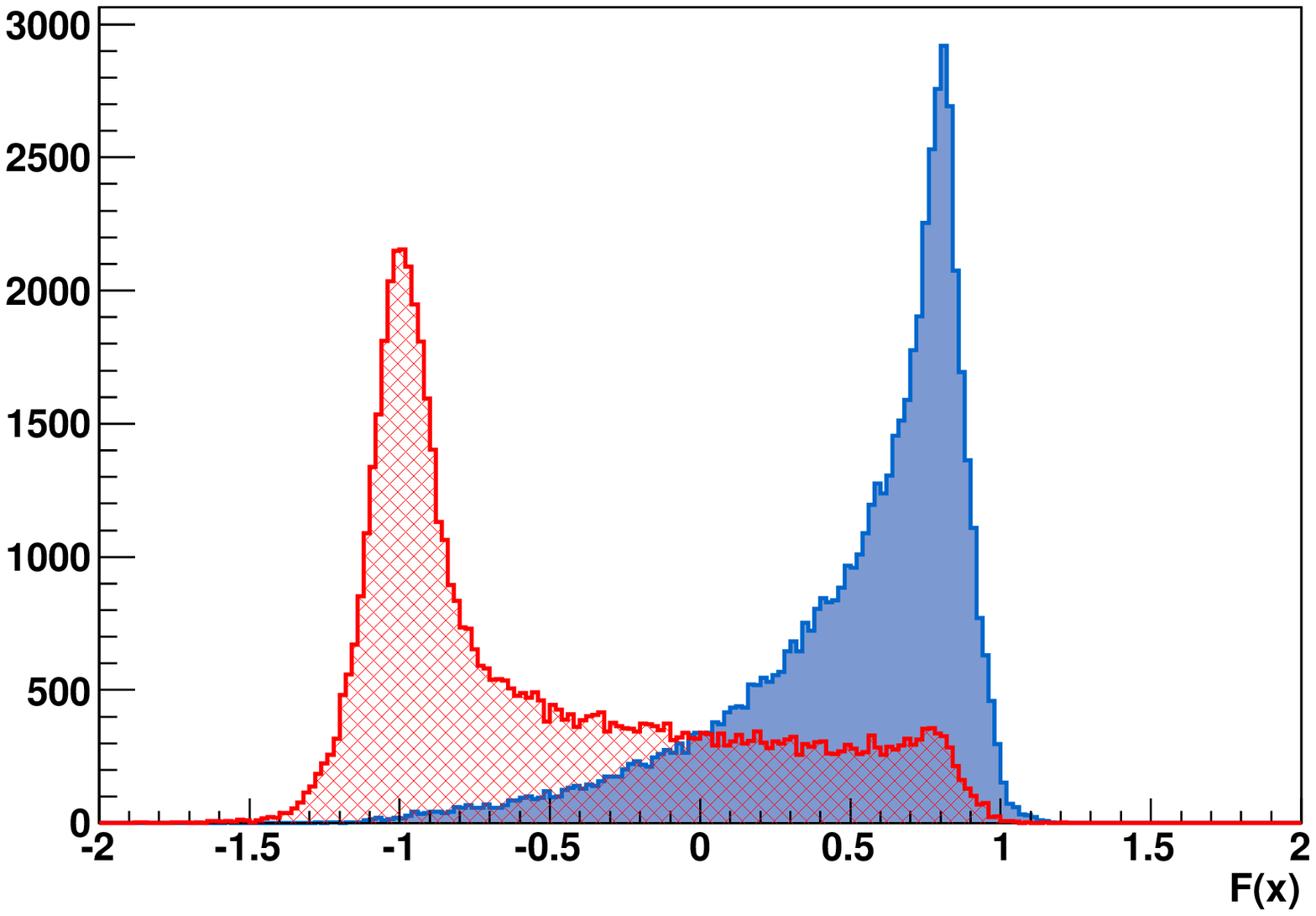}  
		\label{fig:MultiResponse}
	}
	\subfloat[monotonic prediction of the target]{
		\includegraphics[width = 3in]{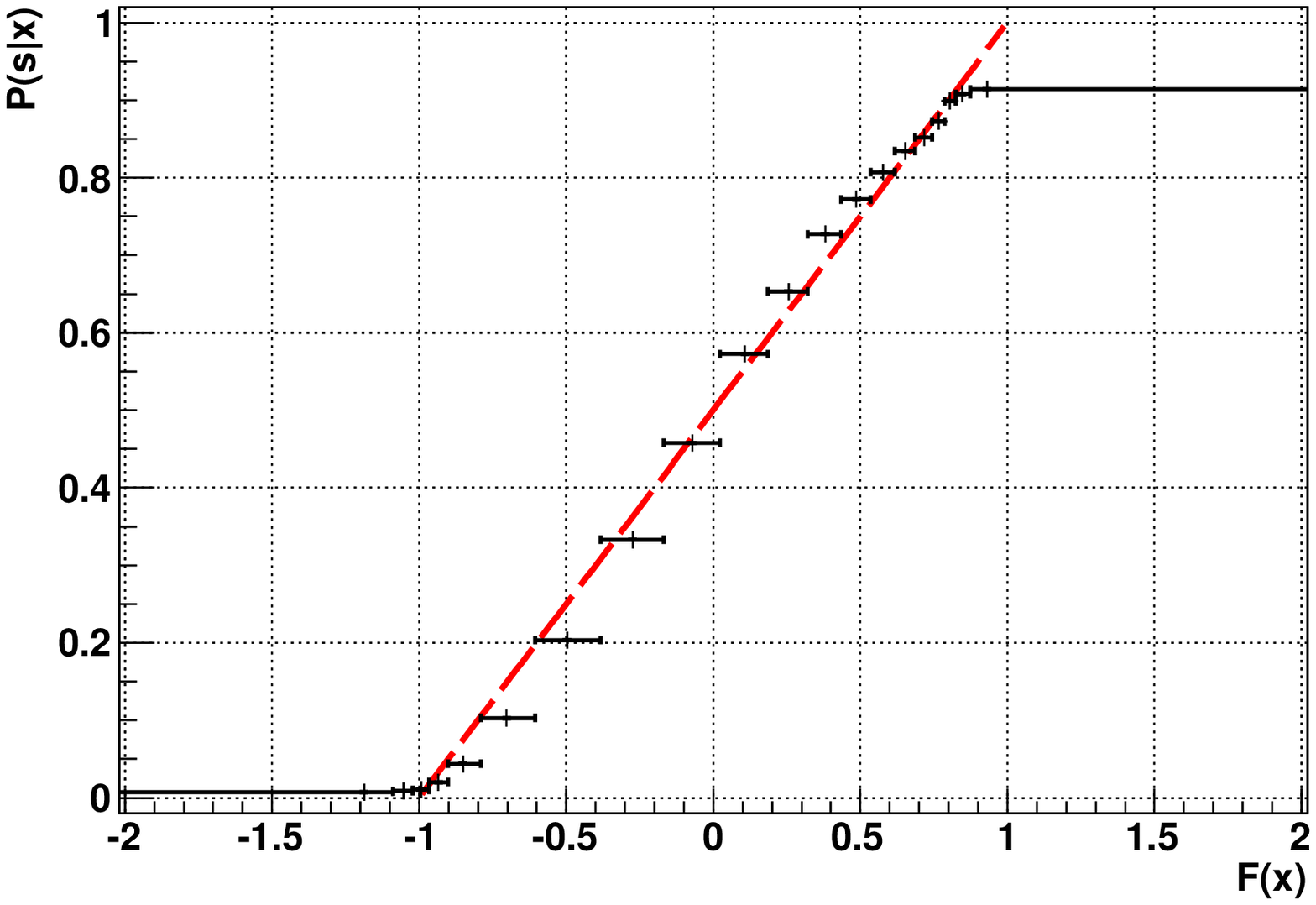}  
		\label{fig:MultiResponseLinearity}	
	}
	\caption{\small \ref{fig:MultiResponse} Separation of signal (solid blue) and background (meshed red) events with the 20 degree multi-polynomial $F(x)$.
		\ref{fig:MultiResponseLinearity} Slightly non-linear, but still monotonic prediction of signal purity with $F(x)$.}
\end{figure}

Figure~\ref{fig:MultiResponse} shows the histograms of the response, and although the $F(x)$ values slightly overshoot~$\pm1$, the response vs. purity on fig.~\ref{fig:MultiResponseLinearity} is still a strictly monotonically ascending curve, assuring that $F(x)$ approximates well the optimal classification contours. For higher dimensional inputs, it is usually enough to approximate the classifier function with a low degree polynomial to have a good estimate on the classification contours, or on the separating power of a new variable. Nevertheless, the method seems to be stable against overfitting, since as it is fed with well determined moments of the distributions, and not with the individual events itself; it is not expected to be sensitive to the high frequency noise associated with sampling. The method is also capable of fitting a non-binary target. In this case the $\hat h^k$ tensors are expectation values of $y$ target multiplied with the $x^k$ moments of the input parameters, $\hat h^k = \avg{y\cdot x^k}$, while $\hat g^k = \avg{x^k}$.

As a remark, it must be noted, that since certain distributions have diverging moments, it is beneficiary to transform the distributions into compact phase spaces prior to training, in order to have evaluable results.

\section*{Acknowlegments}

I would like to acknowledge the support of my colleges, particularly A. Elizabeth Nuncio Quiroz, Ian C. Brock and Eckhard von T\"orne.
\nocite{refROOT}

\bibliography{BibTex/references.bib}

\begin{thebibliography}{5}
\providecommand{\natexlab}[1]{#1}
\providecommand{\url}[1]{\texttt{#1}}
\expandafter\ifx\csname urlstyle\endcsname\relax
  \providecommand{\doi}[1]{doi: #1}\else
  \providecommand{\doi}{doi: \begingroup \urlstyle{rm}\Url}\fi

\bibitem[Anderson et~al.(1999)Anderson, Bai, Bischof, Blackford, Demmel,
  Dongarra, Du~Croz, Greenbaum, Hammarling, McKenney, and Sorensen]{laug}
E.~Anderson, Z.~Bai, C.~Bischof, S.~Blackford, J.~Demmel, J.~Dongarra,
  J.~Du~Croz, A.~Greenbaum, S.~Hammarling, A.~McKenney, and D.~Sorensen.
\newblock \emph{{LAPACK} Users' Guide}.
\newblock Society for Industrial and Applied Mathematics, Philadelphia, PA,
  third edition, 1999.
\newblock ISBN 0-89871-447-8 (paperback).

\bibitem[Ballard et~al.(2011)Ballard, Kolda, and Plantenga]{Ballard6008988}
G.~Ballard, T.~Kolda, and T.~Plantenga.
\newblock Efficiently computing tensor eigenvalues on a gpu.
\newblock In \emph{Parallel and Distributed Processing Workshops and Phd Forum
  (IPDPSW), 2011 IEEE International Symposium on}, pages 1340 --1348, may 2011.
\newblock \doi{10.1109/IPDPS.2011.287}.

\bibitem[Bishop(2006)]{Bishop:2006:PRM:1162264}
Christopher~M. Bishop.
\newblock \emph{Pattern Recognition and Machine Learning}.
\newblock Springer-Verlag New York, Inc., Secaucus, NJ, USA, 2006.
\newblock ISBN 0387310738.
\newblock ISBN 0-387-31073-8.

\bibitem[Brun and Rademakers(1996)]{refROOT}
Rene Brun and Fons Rademakers.
\newblock Root - an object oriented data analysis framework.
\newblock In \emph{Proceedings AIHENP'96 Workshop, Lausanne}, volume 389
  (1997), pages 81--86. Nucl. Inst. \& Meth. in Phys. Res. A, Sep. 1996.
\newblock See also http://root.cern.ch/.

\bibitem[Neyman and Pearson(1933)]{Neyman01011933}
J.~Neyman and E.~S. Pearson.
\newblock On the problem of the most efficient tests of statistical hypotheses.
\newblock \emph{Philosophical Transactions of the Royal Society of London.
  Series {\rm A}}, 231\penalty0 (694-706):\penalty0 289--337, 1933.
\newblock \doi{10.1098/rsta.1933.0009}.
\newblock URL
  \url{http://rsta.royalsocietypublishing.org/content/231/694-706/289.short}.

\end{thebibliography}

\end{document}